\documentclass[letterpaper, 10 pt, conference]{ieeeconf}  % Comment this line out if you need a4paper	

\IEEEoverridecommandlockouts                              % This command is only needed if 	
\usepackage{graphics} % for pdf, bitmapped graphics files	
\usepackage{epsfig} % for postscript graphics files	
\usepackage{amsmath} % assumes amsmath package installed	
\usepackage{amssymb}  % assumes amsmath package installed	
\usepackage{hyperref}	
\usepackage[per=slash]{siunitx}	
\usepackage[mode=buildnew]{standalone} % tikz standalone figures	
\usepackage{subfigure}  % subfigure usage	
\usepackage{booktabs}  % fancy tables	
\usepackage{graphicx}

\usepackage[per=slash]{siunitx}
\usepackage{comment}
\usepackage[normalem]{ulem}
\usepackage{amsmath}
\usepackage{xcolor}
\usepackage{xspace}

\definecolor{OliveGreen}{RGB}{0,200,25}
\newcommand{\red}[1]{\textcolor{red}{#1}}

\newcommand{\darkgreen}[1]{\textcolor{OliveGreen}{#1}}

\newcommand{\eg}{e.\,g.\ }

\newcommand{\softhand}{\mbox{Finger-Vision Soft Hand}\xspace}

\newcommand{\ackInopro}{This work has been supported by the German Federal Ministry of Education and Research (BMBF) under the project INOPRO (16SV7665).}

\newcommand{\replaced}[2]{\red{\ifmmode\text{\sout{\ensuremath{#1}}}\else\sout{#1}\fi}\darkgreen{#2}}
\newcommand{\removed}[1]{\red{\ifmmode\text{\sout{\ensuremath{#1}}}\else\sout{#1}\fi}}

\newcommand{\removedfootnote}[1]{\footnote{\removed{#1}}}
\newcommand{\removedsubsection}[1]{\subsection{\texorpdfstring{\removed{#1}}{#1}}}

\title{\LARGE \bf	
	A Soft Humanoid Hand with In-Finger Visual Perception	
}

\author{Felix Hundhausen, Julia Starke and Tamim Asfour% <-this % stops a space	
	\thanks{\ackInopro}%	
	\thanks{The authors are with the Institute for Anthropomatics and Robotics, Karlsruhe Institute of Technology, Karlsruhe, Germany. {\tt \{felix.hundhausen, asfour\}@kit.edu}}%	
}

\begin{document}

\maketitle	
\thispagestyle{empty}	
\pagestyle{empty}

% CONTENT %%%%%%%%%%%%%%%%%%%%%%%%%%%%%%%%%%%%%%%%%%%%%%%%%%%%%%%%%%%%%%%%%%%%%%%%	
%Proposing a new hardwaresetup for a five finger hand: Humanoid soft finger hand with integration of miniature cameras in fingertips: finger-tip-vision (FTV)	
%State of the art: other type of pre touch information used, cameras attached to endeffector, jaw gripper with 2 cameras, not in humanoid 5 finger hand	
%Goal: Close loop in reaching phase with in finger visual information	
%Proposal of the desing of a  soft finger hand wiht integrated actuation and information processing system	
%Durability test of mechanic/electrical structure	
%Experiment: Localize object during grasps in in finger camera image	
%Evaluation of object segmentation accuracy	

%propose new hardware concept: Finger-Tip VIsion, evaluate with prototype: Soft finger vision hand	

%%%%%%%%%%%%%%%%%%%%%%%%%%%%%%%%%%%%%%%%%%%%%%%%%%%%%%%%%%%%%%%%%%%%%%%%%%%%%%%%	
	
\begin{abstract}	
%The expectations of humanoids hands to mimic the human hands capabilities is a largely still unsolved challenge in robot design. While it is impossible to recreated equal sensor and actuator performance similar humans, alternative and alternated mechatronic setups need to be explored, to realize robotic hands with human-like capabilities as expected and required for service robots but also in use of prosthetic hands. This gives the motivation to find solutions for improved hardware setups with today's available technology and justifiably costs.	

We present a novel underactued humanoid five finger soft hand, the KIT \softhand, which is equipped with cameras in the fingertips and integrates a high performance embedded system for visual processing and control.  We describe the actuation mechanism of the hand and the tendon-driven soft finger design with internally routed high-bandwidth flat-flex cables. For efficient on-board parallel processing of visual data from the cameras in each fingertip, we present a hybrid embedded architecture consisting of a field programmable logic array (FPGA) and a microcontroller that allows the realization of visual object segmentation based on convolutional neural networks. 	
We evaluate the hand design by conducting durability experiments with one finger and quantify the grasp performance in terms of grasping force, speed and grasp success. The results show that the hand exhibits a grasp force of \SI[multi-part-units=single]{31.8(12)}{\newton} and a mechanical durability of the finger of more than $15.000$ closing cycles. Finally, we evaluate the accuracy of visual object segmentation during the different phases of the grasping process using five different objects. Hereby, an accuracy above \SI{90}{\percent} can be achieved.

\end{abstract}

%%%%%%%%%%%%%%%%%%%%%%%%%%%%%%%%%%%%%%%%%%%%%%%%%%%%%%%%%%%%%%%%%%%%%%%%%%%%%%%%	

\section{Introduction}	
	
%\begin{itemize}	
%	
%\item progress in vision based grasping based on learning and cnns (mention this?)	
%	
%\item typically used hardware setups (eye-in-hand, head mounted external)	
%	
%\item our new setup: camera in fingertip	
%	
%\item enabled by technology miniaturization	
%	
%\item this work: design a system (why soft fingers: low cost for complex mechanical structure, high adaptability), evaluate possibilities	
%\end{itemize}	
	
The design of robotic hand comprises the challenges of designing an actuation system as well as a sensor system that can provide full-featured feedback for a controller that generates actuation signals. 	
Visual sensor systems are widely used in robotics and provide, not only due to recent progress in deep learning based vision methods, valuable information about the scene in which a robot perform its tasks.	
Compared to an external camera, an end-effector-mounted camera allows to minimize the position error independently of the limited accuracy of the robot kinematics \cite{hutchinson1996tutorial}.

\begin{figure}[t!]	
	\centering	
	\includegraphics[width=1\linewidth]{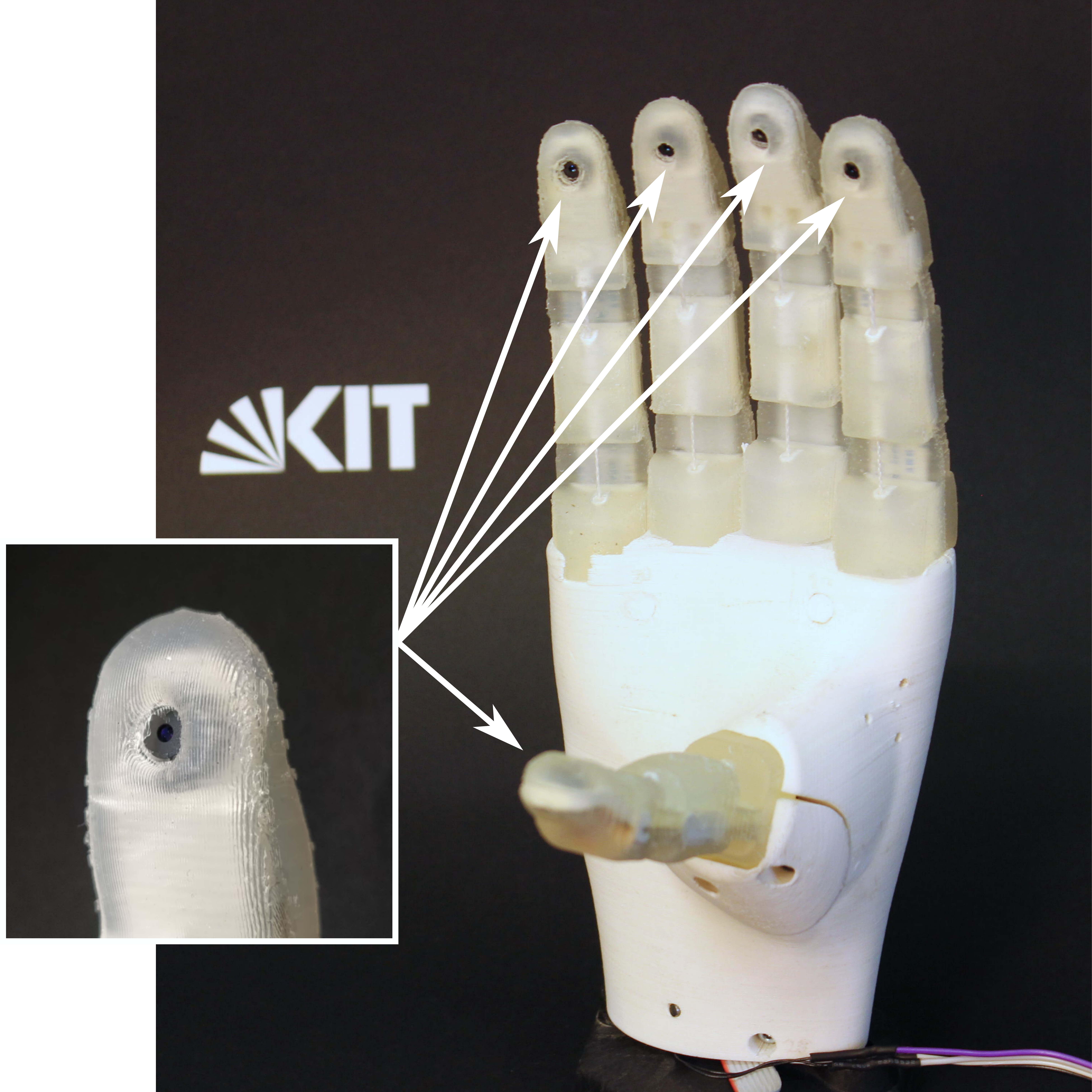}	
	\caption{The \textit{\softhand} with 2~Megapixel in-finger cameras.}	
	\label{fig:model}	
\end{figure}

In this work, we present the design of the KIT \softhand, as shown in Fig. \ref{fig:model}, that includes cameras directly inside the tips of soft fingers. The use of such in-finger cameras is enabled by their miniaturization, driven by the demand for smartphones and laptops. The integration of cameras inside a robot fingertip provides redundant visual information without inevitable occlusion by the gripper itself before establishing contact with the objects to be grasped, and provides the advantages of multi-camera based perception. Further, pose recovery for the cameras allows inferring finger poses without internal sensors, a promising approach that can contribute to the estimation of soft robotic structures such as the finger of the hand described in this work.	

The paper is structured as follows: In section \ref{sec:relWork} we give an overview of related work regarding optical methods for in-finger perception and describe relevant soft fingers found in literature. In section \ref{sec:design} we present the design of the \softhand that includes the design of soft fingers with in-finger cameras and data connection. Further we present integrated actuation mechanic and a hybrid embedded system for data processing and control. Subsequently we propose a convolutional encoder-decoder network architecture for the extraction of semantic data from the visual data. We evaluate the system performance in section \ref{sec:eval} including grasp performance, mechanical durability of the finger design and evaluation of the perception system during a grasping experiment.

\section{Related Work}	
%\textbf{Optical environment perception for huamanoids hands:}	
%\textbf{In finger located vision based perception}	

\label{sec:relWork}
In the development of robotic hands, soft mechanisms such as soft fingers with flexible joints receive an increasing attention, as they facilitate a safe and compliant adaptation of the hand towards the objects to be grasped, while providing mechanical robustness and increasing grasp stability~\cite{piazza2019century}.	
While soft robotic grippers are designed with a wide variety of shapes and actuation principles \cite{Rus2015,Laschi2016}, an increasing number of soft humanoid hands has been presented throughout the recent decade.

Several humanoid hands have been presented including pneumatically actuated soft fingers with rigid \cite{Gaiser2008} and continuous joints \cite{Deimel2016,Tian2017}.
Continuous designs exhibit a highly compliant object interaction, while their grasping strategies differ from the human hand due to their continuous joint structure.	
Alternatively, rigid finger segments can be combined with flexible joints, which can adapt to external influences.
The four-fingered iHY hand uses shape deposition modeling to manufacture elastic joints \cite{odhner2014compliant}.
Other design approaches for flexible joints include compliant rolling contact joints \cite{Catalano2014} and compliant spring joints integrated into the structure of a monolithic finger \cite{Melchiorri2013}.	

Besides the design of soft flexible fingers, their sensorization is an other key aspect for successful grasping and manipulation and multiple sensor setups and modalities have been applied in robotic hands \cite{saudabayev2015sensors}.	
Both, internal and external information can be gathered by visual sensor setups.	
Interoceptive visual sensor systems are used as tactile sensors to detect forces and torques by perceiving the deforming shape of an elastic surface. Discrete photoreceivers can detect changing optical characteristics \cite{hasegawa2010development}. Also, camera systems including optical lenses can be used to track internal features on the finger surface (\cite{knoop2013dual,li2014GelSight,zhang2019vision}) that allow reconstruction of the outer shape of the fingers. A review of camera-based methods can be found in Shimonomura et al. \cite{shimonomura2019tactile}.

Furthermore, optical sensor setups can also be applied to directly detect exteroceptive information before establishing contact. %Active sensors emit light from the finger-structure and then detect reflected light to infer information regarding the distance of an object.
Active proximity sensors that emit and detect infrared light have been \eg integrated in robotic grippers (\cite{balek1985using}, \cite{yamaguchi2018gripper}) and a three-fingered hand \cite{hsiao2009reactive}.

The principle of using cameras \textit{inside the fingertips} to obtain visual pre-touch information is barely investigated in existing literature. Robotic grippers that use cameras inside the fingers for exteroceptive sensing can be found in \cite{yamaguchi2016} and \cite{shimonomura2016robotic}.	
Shimonomura et al.~\cite{shimonomura2016robotic} use a stereo camera system inside a parallel jaw gripper to obtain depth/proximity information. Additionally touch information is provided by an infrared vision system and a light conductive plate in front of the camera. This enables to execute tasks like searching, approaching and grasping an object.	
Yamaguchi et al. presents a system that combines vision and marker based tactile feedback \cite{yamaguchi2016}. Markers on a translucent elastic surface are tracked by a finger-integrated camera, that can provide forces and torques asserted on the finger.	
To our best knowledge, no humanoid five-finger hand with integrated exteroceptive visual perception in each finger has been presented in the literature yet.

The advantage of visual feedback from an in-hand integrated camera was investigated in our previous work~\cite{Weiner2018} and \cite{hundhausenresource} with the objective of the development of a context-aware prosthetic hand.
Kinematic control strategies based on gripper-mounted camera was shown in multiple visual servoing based approaches \cite{lampe2013acquiring}, in learning based approaches (\cite{yan2017simtoreal},) or in a combination of both  (\cite{viereck2017learning}, \cite{morrison2018closing}.

\section{The \softhand}
\label{sec:design} 

The mechanical basic structure of the hand is based on our previous work \cite{Weiner2018}, from which we adopt the tendon based underactuation scheme.
The \softhand includes five soft fingers, equipped with visual sensors, that are adaptively underactuated by three motors.
For the control of the hand, we present an embedded controller board specifically designed for the requirements resulting from our hardware and sensor setup. It includes a high performance real-time data-processing system as well as motor control circuits, power management and required communication interfaces.
We present a network architecture for visual object segmentation designed for real-time inference.
The dimensions of the hand are designed to match the human size while integrating all motors and gears, the underactuation mechanism and the controller board. The shape of the hand palm was derived from a CAD model~\cite{grabcad} of a human hand as a reference.
The final prototype results in a total weight of \SI{580}{\gram} and \SI{28.6}{\gram} per finger. All fingers have a length of \SI{10}{\centi\meter} and width of \SI{1.7}{\centi\meter} for the intermediate joint. The total hand length is \SI{21.5}{\centi\meter}.

\begin{figure}[b]
	\centering
	%\footnotesize
	%\def\svgwidth{0.5\linewidth}
	%\input{figures/mechanism5F.pdf_tex}
	\includegraphics[width=0.5\linewidth]{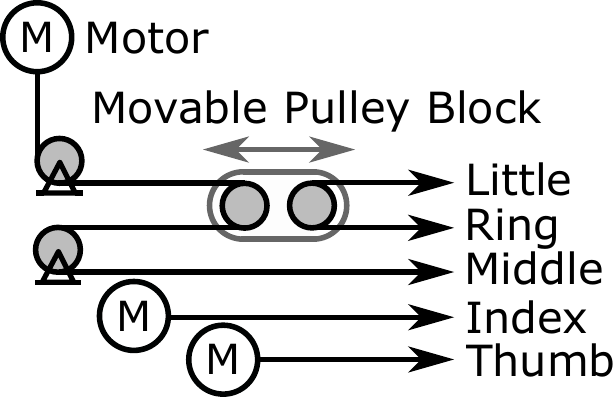}
	\caption{Tendon-based actuation scheme of the hand: Two separate motors are used for actuation of the index finger and thumb. A third motor actuates middle, ring and little finger using an adaptive mechanism for equal force distribution.}
	\label{fig:mechanism}
\end{figure} 

\subsection{Mechanical design}

The hand consists of a rigid palm which is FDM-printed using ABS and five tendon actuated silicone-casted soft fingers, which include the cameras and flat-flex-cables for electrical interconnection.
To allow the realization of a variety of grasps including precision and power grasp types, we include three motor gear units of type Faulhaber 2224U012SR 20/1R with a planetary gear with a transmission factor of 23:1 inside the palm. The motor's angular velocity is measured using relative encoders (Faulhaber IEH2-512), that are attached to the motor-shaft and provide a resolution of 512 impulses per revolution. The thumb and index fingers are directly driven by tendons (Dyneema, \SI{0.4}{\milli\meter}) reeled up on pulleys on the gear shaft. The middle, ring and little finger are jointly actuated using an adaptive underactuated mechanism that equally distributes the force from one actor to multiple fingers. This reduces the complexity of control and mechanical design.
The underactuation mechanism (see Fig. \ref{fig:mechanism}) is a modified version of the 50\textsuperscript{th} percentile Female KIT Prosthetic Hand mechanism (\cite{Weiner2018}), which is based on the TUAT-Karlsruhe mechanism~\cite{fukaya2000design}. Instead of using only two actuators, the \softhand presented in this paper contains an additional third actuator.  

The tendon coming from the motor assigned to the coupled little, middle and ring finger is routed over a roller of a movable pulley block and ends at the third finger. The other fingers are connected over the second roller of the pulley block. The complete actuation scheme is depicted in Fig. \ref{fig:mechanism}. If friction is neglected, the mechanism distributes the force to all three fingers equally. For routing of the tendons in between motors, the mechanism and the fingers, we use a combination of two rollers and low-friction PTFE-tubes embedded into the 3D-printed hand structure.

\begin{figure}[t]
	\centering
	\includegraphics[width=1\linewidth]{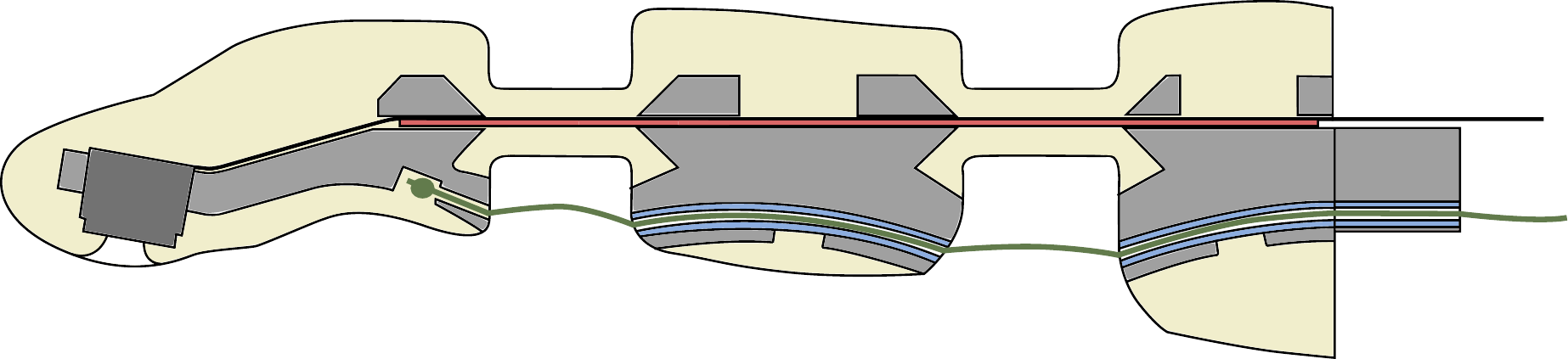}
	\caption{Cut view of the silicone-material (colored in beige) enclosing the rigid bone segments (gray) and reinforcing PET-strip (red) which is glued to the rigid segments on the bottom side. The tendon (green) is guided trough PTFE-tubes (blue) in the rigid segments. Camera: colored dark gray}
	\label{fig:fingerCutView}
\end{figure}
The finger kinematic consist of two flexible joint that represent  the metacarpophalangeal and proximal interphalangeal joint of the human finger. A cross section of the finger structure is shown in Fig. \ref{fig:fingerCutView}. The joint structure is adapted from the iHY-finger presented in \cite{odhner2014compliant}. However, our setup includes two flexible joints instead of one.
The fingers rigid bone segments are interlinked by a flexible PET strip and are enclosed in casted silicone (45~ShA). The monolithic soft material functions as an elastic interconnection between the finger segments. In addition it mimics the soft tissue of the human finger and provides a deformable, high friction material for interaction with objects.

\begin{figure}[tb]
	\centering
		\subfigure[]{
		\parbox{0.225\textwidth}{
			\label{fig:bs1}
			\includegraphics[trim={0cm 0 0cm 0cm},clip, width=0.22\textwidth]{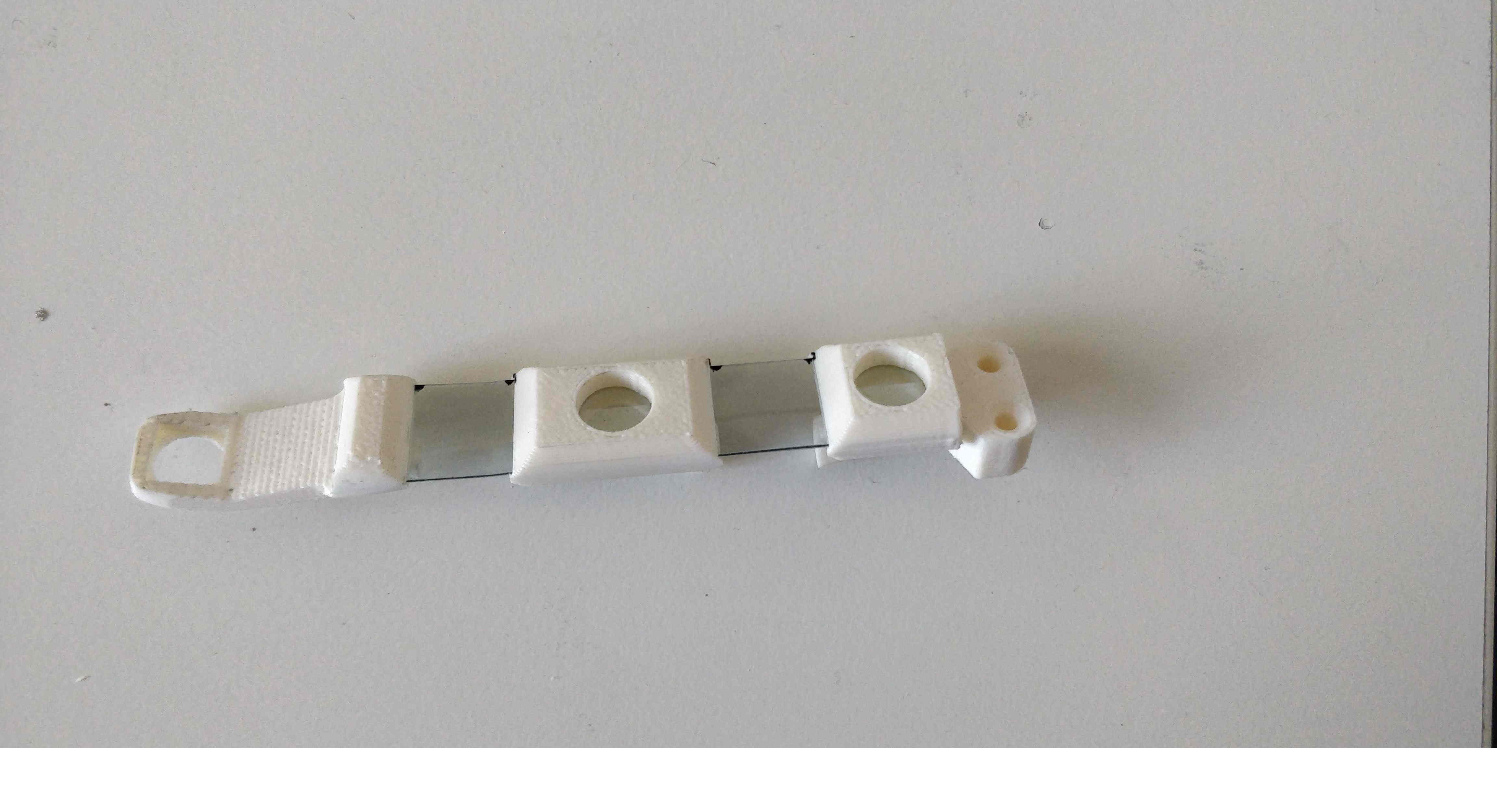}
	}}
	\subfigure[]{
		\parbox{0.225\textwidth}{
			\label{fig:bs2}
			\includegraphics[trim={0cm 0 0cm 0cm},clip, width=0.22\textwidth]{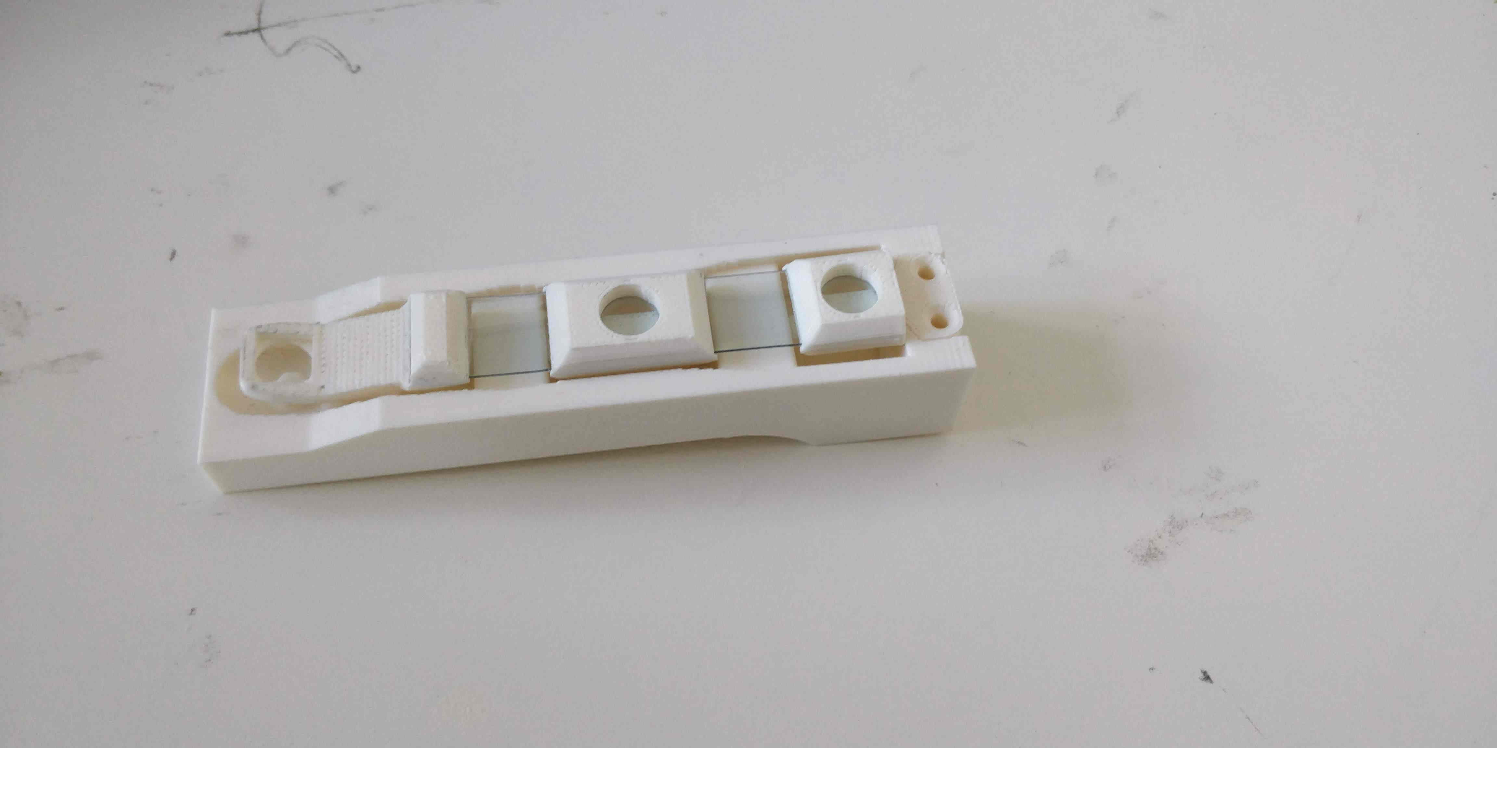}
	}}
	
	\subfigure[]{
		\parbox{0.225\textwidth}{
			\label{fig:bs3}
			\includegraphics[trim={0cm 0 0cm 0cm},clip, width=0.22\textwidth]{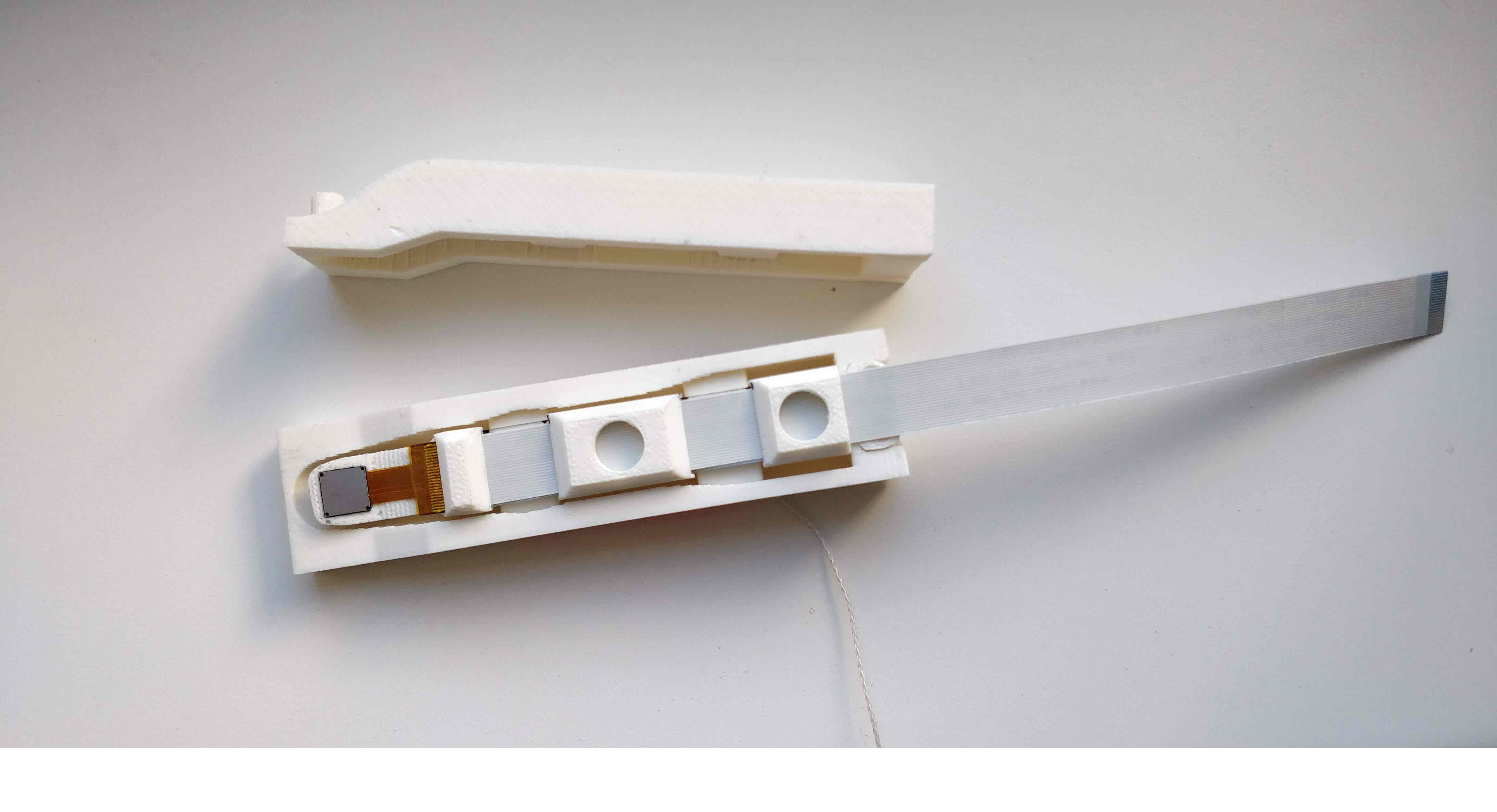}
	}}
	\subfigure[]{
		\parbox{0.225\textwidth}{
			\label{fig:bs4}
			\includegraphics[trim={0cm 0 0cm 0cm},clip, width=0.22\textwidth]{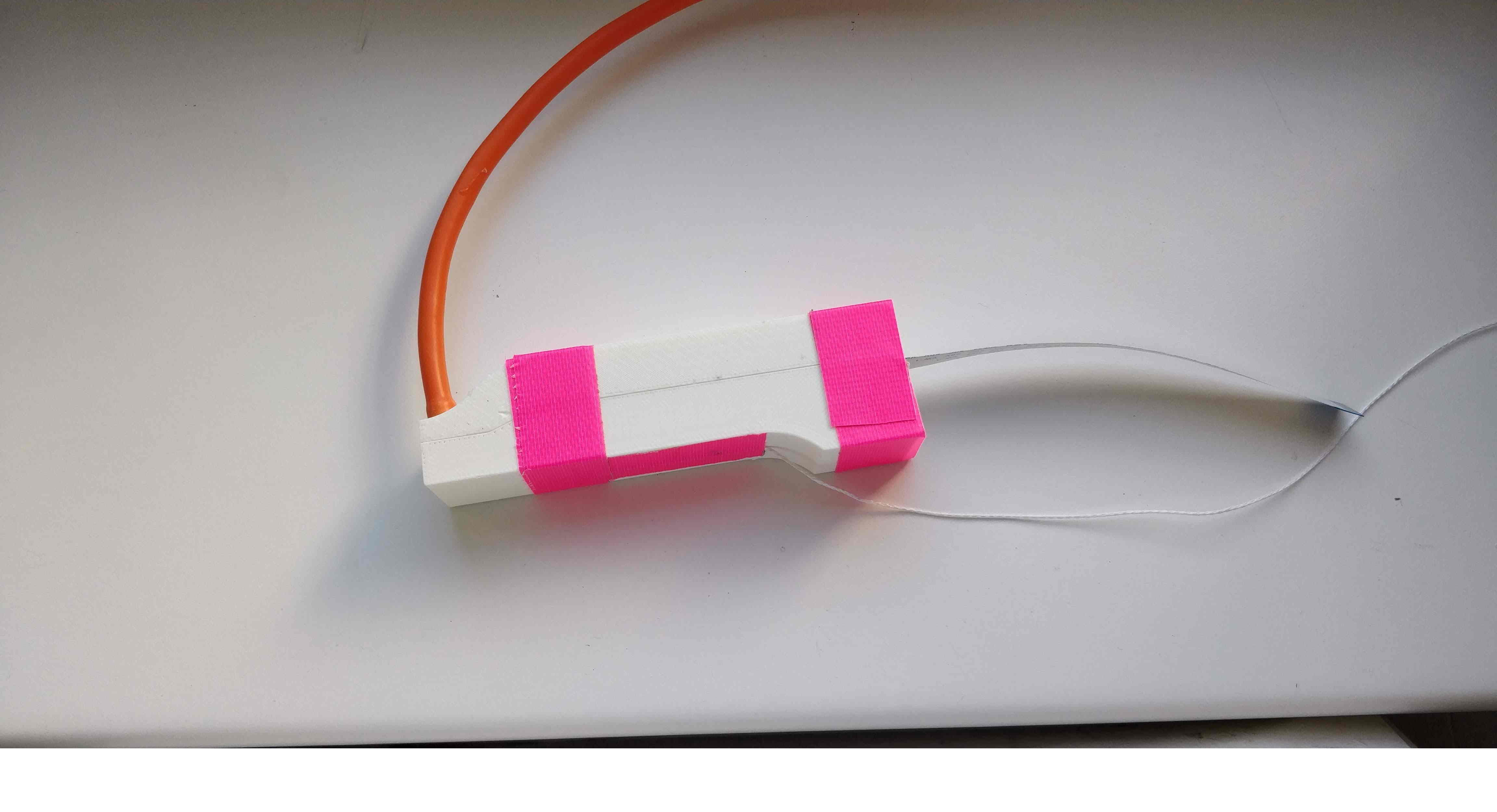}
	}}

	\subfigure[]{
		\parbox{0.225\textwidth}{
			\label{fig:bs5}
			\includegraphics[trim={0cm 0 0cm 0cm},clip, width=0.22\textwidth]{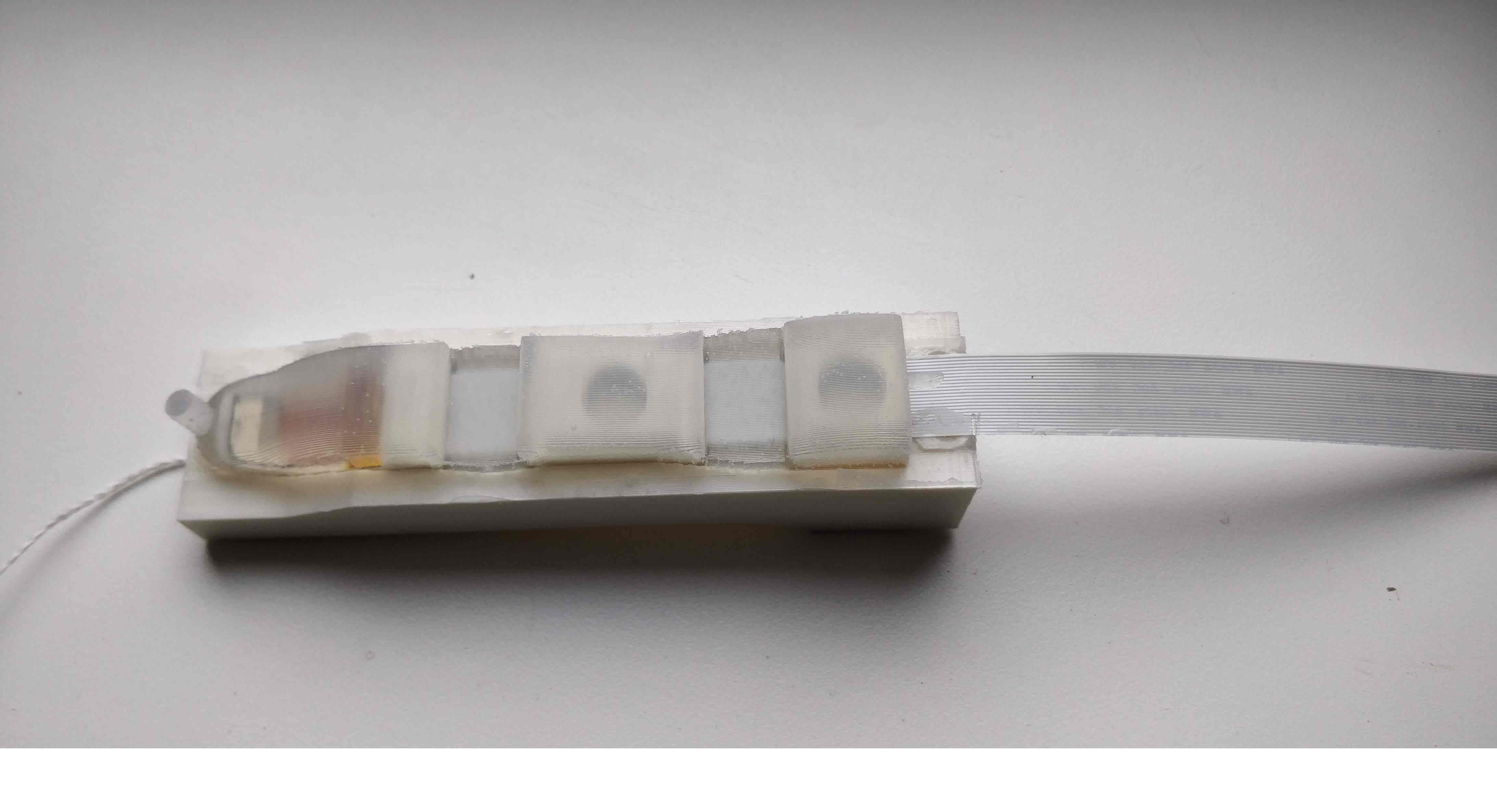}
	}}
	\subfigure[]{
		\parbox{0.225\textwidth}{
			\label{fig:meanBs}
			\includegraphics[trim={0cm 0 0cm 0cm},clip, width=0.22\textwidth]{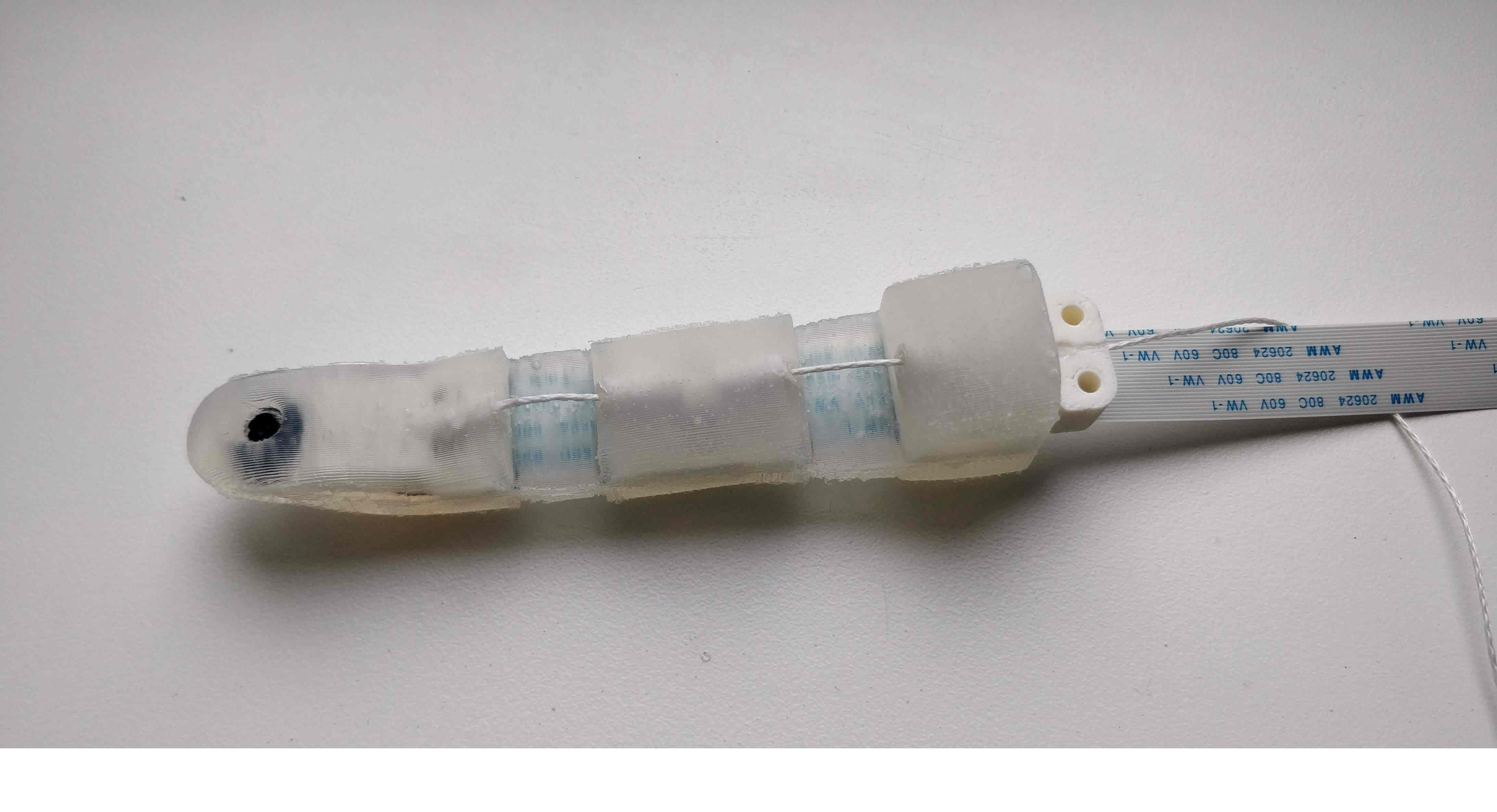}
	}}
	
	\caption{Manufacturing steps of the soft fingers: Rigid bone segments connected by a PET-strip for reinforcement (a) are placed in a two part mold (b). The camera is inserted into the distal bone segment and is connected by a flat-flex cable that follows the PET-strip (c). The mold is closed and a tube attached, through which the mold is filled with silicone. After the silicone is cured, the mold is opened (e) and the finger is taken out. The silicone from inside the tendon guiding tubes and from the filling channels is removed and the tendon is inserted. Also, the silicone residues and the protection film in front of the camera lens are removed (f).
	}
	\label{fig:art}
	
\end{figure}

The tendon routing channels are realized by PTFE-tube segments glued into the rigid bone segments.
The camera cable (24 pin, 0.5~mm pitch) follows the back side of the PET-strip that interconnects all rigid finger bone segments. This interconnection prevents pulling forces on the cable and acts as a neutral bending axis inside the elastic structure that does not experience length changes during flexion. The manufacturing process using a mold, in which the inner finger structure is placed and casted with two components silicone, is illustrated in Fig. \ref{fig:art}.

\subsection{Embedded Electronic System}
\begin{figure}[tb]
	\centering

	\subfigure[]{
		\parbox{0.48\textwidth}{
			\label{fig:bd1}
			\includegraphics[trim={0cm 0 0cm 0cm},clip, width=0.47\textwidth]{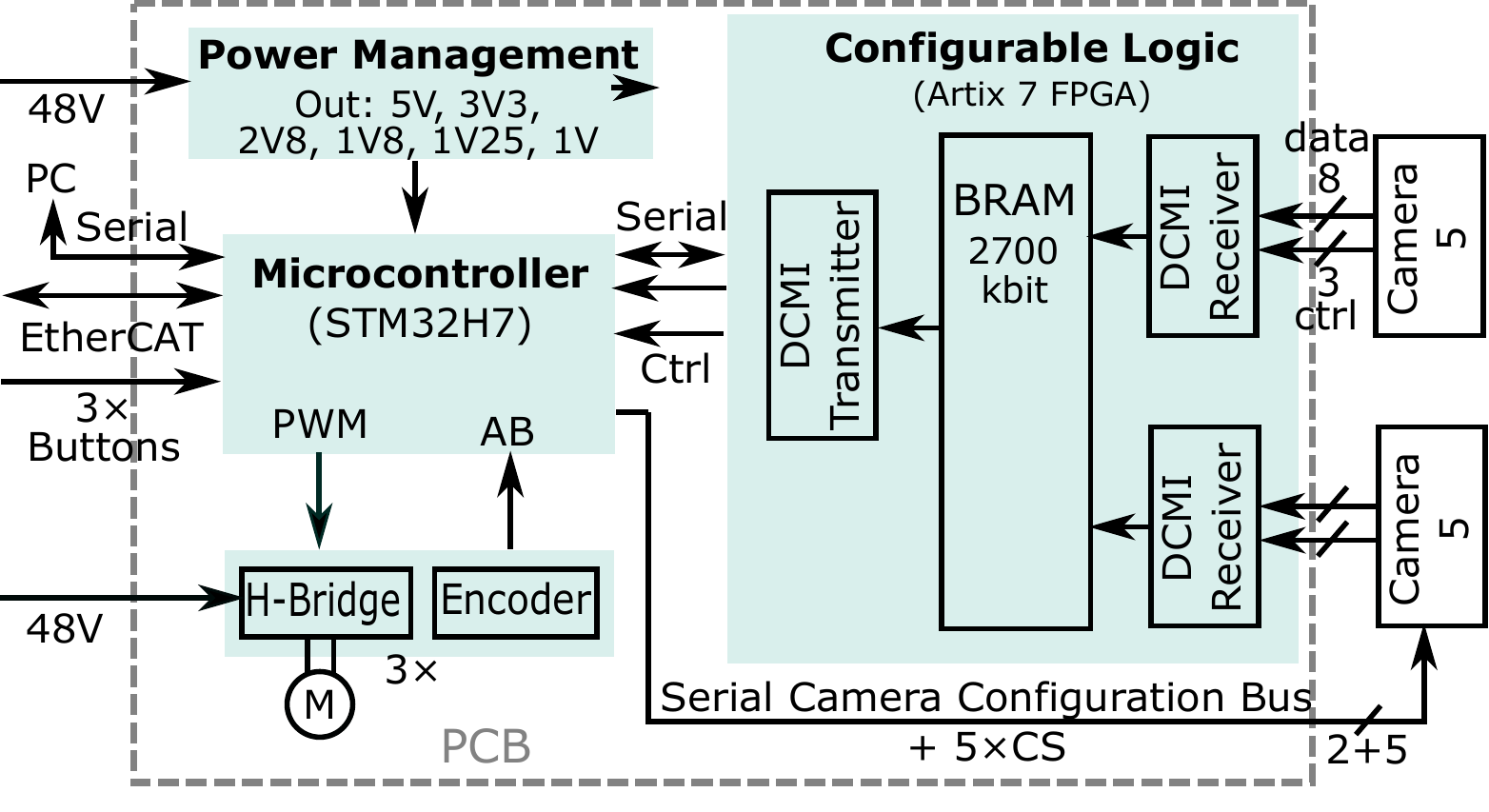}
			\label{fig:blockdiagramm}
	}}
	\subfigure[]{
		\parbox{0.46\textwidth}{
			\label{fig:pcb}
			\includegraphics[trim={0cm 0cm 0cm 0cm},clip, width=0.45\textwidth]{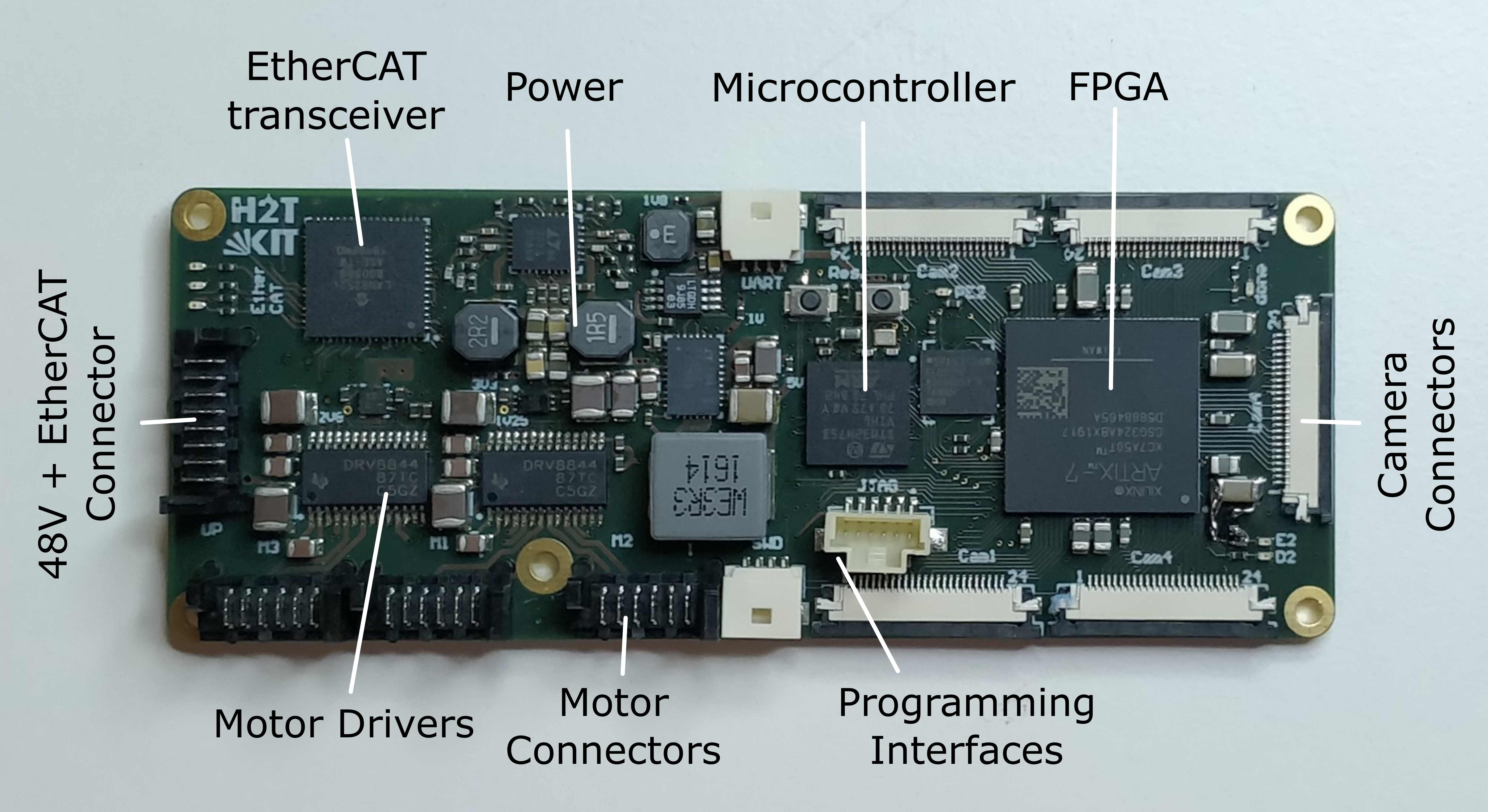}
			\label{fig:PCB}
	}}
	\caption{Hand internal controller PCB. a) Block Diagram b) Photo of the PCB used for experiments with relevant components labeled}
\end{figure}

Since grasping requires real-time processing of the data, the performance of the data-processing architecture is crucial for responsiveness of the hand. To allow parallel receiving and processing of image data from the multiple cameras, that each provides a high data rate signal, we address the challenge of efficient data transfer and processing hardware. Compared to most available microcontrollers, which are limited in number of high bandwidth camera interfaces, reconfigurable logic such as FPGAs allows configuration of parallel data processing structures and provides sufficient IO-pins. This make FPGAs suitable for the underlying hand with its multiple in-finger cameras. For tasks like motor control and lower data-rate interfaces, microcontrollers provide advantages of procedural programming methods, hence we choose an additional microprocessor, that supplements the FPGA. The resulting hybrid architecture consisting of an FPGA and a microcontroller is depicted in Fig. \ref{fig:blockdiagramm}. It allows parallel data processing of visual data as well as procedural program control using the processor sub-system.
The embedded system further includes three DC-motor drivers, an EtherCAT real-time bus interface, as well as a set of voltage converters to provide the required supply voltages for connected sensors and the data-processing components. The system tolerates input voltages ranging from 24~V to 48~V DC, intended for compatibility with typical robot supply rails.
The complete system was realized on a \SI{90}{\milli\meter} by \SI{35}{\milli\meter} PCB as shown in Fig. \ref{fig:PCB} that allows integration into human sized robotic hands.
As an FPGA, a 52~k logic cell and 330~kByte block RAM XILINX Artix 7 was selected. The microcontroller is based on an Arm Cortex H7 processor with a clock frequency of 400~MHz that provides 2~MB of flash memory and 1~MB RAM (STMicroelectronics STM32H7).

As shown in Fig. \ref{fig:blockdiagramm}, all cameras (OmniVision OV2640) are connected by 8 bit wide DCMI buses (Digital Camera Interface) and additional control signals directly to the FPGA, which allows to receive data of five parallel streams with 20 frames per second at a resolution of 176 $\times$ 144 RGB (QCIF). The FPGA implements a receiver component that transfers raw camera data into the block RAM (BRAM). The buffered data stream can be further processed by the FPGA-implemented processing units. For experiments conducted in this work, the FPGA is used only for serialization of the multiple in-finger camera data streams for combined forwarding to the processor unit via a single parallel 100~Mbit bandwidth bus following the DCMI protocol. The interface for data transfer from processor to FPGA is realized as a 12.5~Mbit serial bus, provided for use in future work.

\subsection{Low-Level Controller}
The local low-level control of the motors is realized by an embedded controller implemented on the microprocessor.
The motor angles are obtained by relative encoders connected to microcontroller internal timer units that allow measurement of the tendon pulley angle with a resolution of $47104$ steps per revolution. Thus, the complete closing of a finger results in $60.000$ steps for the individually actuated fingers and $180.000$ steps for the fingers coupled by the underactuated mechanism.
The motor voltage is controlled using pulse-width modulation (PWM) with a resolution of $\frac{1 }{3000}$ that is provided by the motor drivers of type Texas Instruments DRV8844. The voltage is controlled by a cascaded velocity and position PID-controller executed in a \SI{1}{\kilo\hertz} control loop.
For hand-guided grasping and manipulation by an operator, a three button interface is provided for motor-wise control.

\subsection{Visual Object Detection}
To demonstrate the possibility of extraction of scene information from the visual data during grasping , we investigate a pixel-wise semantic segmentation of the camera images. Therefore, we implement a convolutional neural network to segment specific objects in the camera image and thereby provide usable scene information for a higher level grasp controller.

The neural network is realized in an encoder-decoder architecture that includes residual connections and a subsequent threshold function to provide a binary pixel-wise classification of the target object in the camera image. The network architecture is adopted from our previous work~\cite{hundhausenresource} and is inspired by work presented in \cite{ronneberger2015unet} and \cite{badrinarayanan2015segnet}. The networks hyperparameters were determined using the evaluation set of our recorded and annotated dataset as described in \ref{subsec:perception}, the final architecture consists of five convolutional layers and one residual connection. All layer types and filter, output and number of operations are listed in Tab. \ref{tab:archi}.
For training, we use binary cross entropy as a loss function and carry out optimization using the Adam method for 150 epochs.

\begin{table}[h!]
	\caption{Network hyperparameters}
	\centering

	\resizebox{\linewidth}{!}{%
		\begin{tabular}{|c|c|c|c|}
			\hline
			Layer Type & Filter Shape & Output Shape & Operations   \\ \hline
			Input & N.A. & $88\times72\times3$ (19.0~kB) & N.A. \\ 
			Convolution & $3\times3\times3\times16$ (432~B) & $88\times72\times16$ (101~kB) & 2.7~M \\ 
			Convolution & $3\times3\times16\times16$ (2.30~kB) & $88\times72\times16$ (101~kB)~\textbf{\textasteriskcentered} & 14.6~M  \\ 
			MaxPooling & N.A. & $22\times18\times16$ (6.34~kB) & N.A. \\ 
			Convolution & $3\times3\times16\times16$ (2.30~kB) & $22\times18\times16$ (6.34~kB) & 0.9~M \\ 
			Upsampling & N.A. & $88\times72\times16$ (101~kB) &  N.A. \\ 
			Concat. with \textbf{\textasteriskcentered} & N.A. & $88\times72\times32$ (203~kB) &  N.A. \\ 
			Convolution & $3\times3\times32\times8$ (2.30~k) & $88\times72\times8$ (50.7~kB) & 14.6~M\\ 
			Convolution & $3\times3\times8\times1$ (72~B) & $88\times72\times1$ (6.34~kB) & 0.5~M  \\ \hline \hline
			Total & 7.4~kB weights & & 33.3~M   \\ \hline
		\end{tabular}%
	}

	\label{tab:archi}
\end{table}

To analyze the performance of real-time inference of the network, we evaluate the number of MAC operations. Layer-wise numbers are included in Tab.~\ref{tab:archi}, the total number for obtaining one output frame results in $33.3$~M operations.
Regarding memory requirements, which is also a limited resource for inference on embedded hardware, we obtain a demand of $7.4$~kB.

\section{Evaluation}

\label{sec:eval}
We evaluate our hand design in terms of grasp performance, mechanical finger durability and visual perception accuracy. The grasp performance is assessed as individual finger forces, total hand force and grasp success on a set of 60 objects. To evaluate the mechanical finger design including electrical connections, we perform a long term durability test with a finger mounted on a test bench setup. We finally evaluate the performance regarding the extraction of semantic information from the in-finger camera images using the described convolutional network architecture.

\subsection{Grasp performance}
To measure grasp forces, we use a calibrated 6DoF force-torque sensor (Mini 40, ATI Industrial Automation) with 15 repetitions per measurement.
To assess individual fingertip forces, the fingers of the open hand are positioned over the sensor coated with high-friction material and the fingers are closed with maximum possible supply voltage.
The resulting fingertip forces range from \SI{6.3}{\newton} for the middle finger to \SI{11.6}{\newton} for the little finger.
Two half cylinders with a diameter of \SI{31}{\milli\meter} are attached on both sides of the force sensor to assess the cylindrical grasp force of the \softhand.
By grasping the half cylinders with a power grasp, the hand achieves a grasp force of \SI[multi-part-units=single]{31.8(12)}{\newton}.
The finger closing speed is extracted from video data in five repetitions.
The hand closing time thereby results to maximal \SI[multi-part-units=single]{1.22(3)}{\second} for the underactuated fingers and \SI[multi-part-units=single]{0.49(3)}{\second} respectively \SI[multi-part-units=single]{0.44(3)}{\second} for thumb and index finger.

To evaluate the grasp functionality of the \softhand, an adapted form of the gripper assessment protocol \cite{doi:10.1177/0278364917700714} is applied. We follow the same evaluation procedure as in our previous work \cite{Weiner2018}.
Contrary to the gripper assessment protocol designed for robot operated hand, the hand is positioned by a human operator and controlled via the three button interface.
We enlarge the used object subset to a total of 60 items from YCB Object Set \cite{doi:10.1177/0278364917700714} that solely excludes the task items.
The objects are lifted from a flat table surface and turned \SI{90}{\degree} within the hand.
Overall, \SI{91.8}{\percent} of the tested objects can be grasped and lifted successfully, resulting in a grasp score of 201.5 out of 230.0 points.

\subsection{Mechanical finger durability test}

\begin{figure}[b]
	\centering
	\includegraphics[width=1\linewidth]{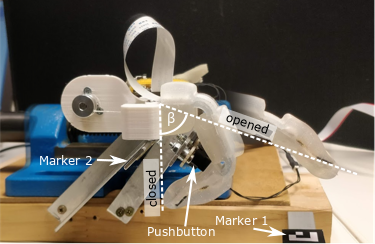}
	\caption{Setup for the longtime durability test with one finger on a test-bench. Opened and closed finger configurations are overlaid.}
	\label{fig:lontimetest}
\end{figure}

Electrical connections in moving robotic segments are prone to failures. This makes the wiring inside the fingers a challenging and crucial task. To evaluate our finger design, we conduct a long time reliability test of the mechanical structure and electrical data connection passing the two soft finger joints.
The long term test is conducted on a newly fabricated finger actuated and controlled similar to hand internal setup on a test bench setup as shown in Fig. \ref{fig:lontimetest}. To detect whether the finger is completely closed, a push button is contacted by the distal finger segment in a closed position. The finger is continuously opened and closed and therefore controlled by a position controller. Two markers are placed at the test bench and are captured by the camera at minimum and maximum finger closing angle $\beta$. This allows evaluation of the recorded images for failed image data and positioning. Additional to the camera images, the contact information from the push button is recorded.

The recorded data is visually checked for corrupted images as well as mechanical finger function.
The first corrupted image was obtained after 4968 finger actuation, which indicates a  failure of an electrical connection of the camera inside the soft finger structure. After this failure, still partially correct image data was obtained, complete signal loss occurred after 5665 actuations.
The mechanical system was fully functional, when the test was terminated after more than 15.000 finger actuations.

\subsection{Visual Perception Experiment}
\label{subsec:perception}

\begin{figure}[tb]
	\centering
	
	\subfigure[]{
		\parbox{0.4\textwidth}{
			\label{fig:experiment}
			\includegraphics[trim={0cm 0cm 0cm 0cm},clip, width=0.4\textwidth]{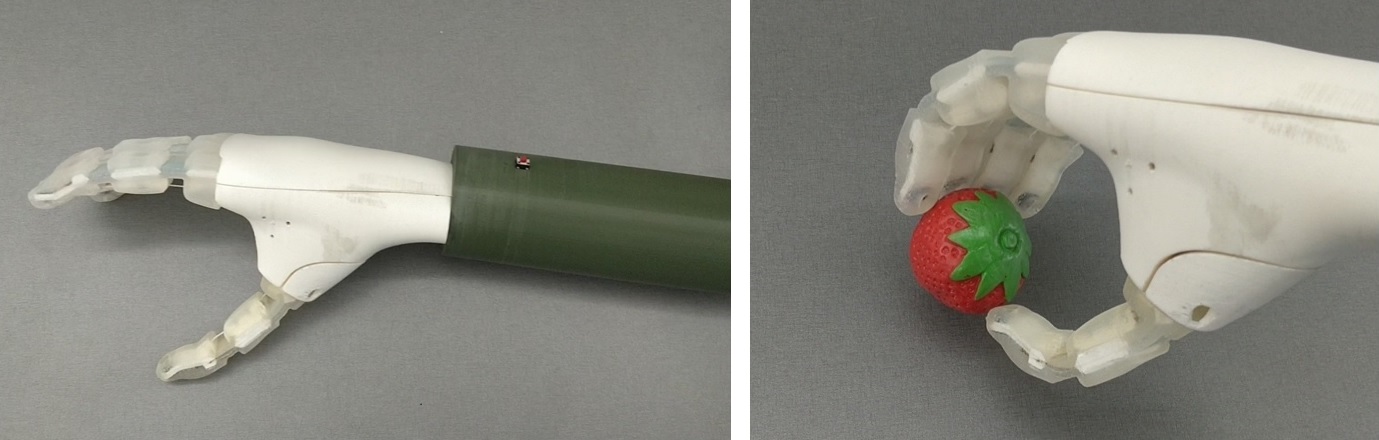}
			\label{fig:exp}
	}}
	\subfigure[]{
		\parbox{0.5\textwidth}{
			\label{fig:segmenation}
			\includegraphics[trim={0cm 0 0cm 0cm},clip, width=0.45\textwidth]{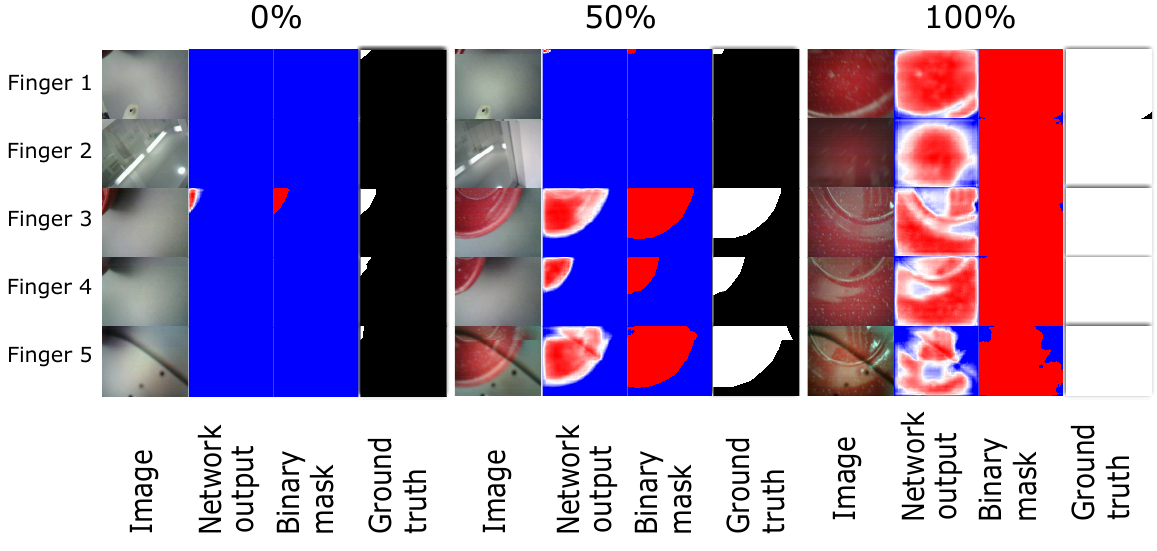}
	}}
	\caption{a) Setup and exemplary grasped object b) Camera stream and segmentation during a grasp experiment with the bowl including the result and ground truth data. The percentage values indicate temporal progress of the grasp execution.}
	\label{fig:graspingexperiment}
\end{figure}

To evaluate the performance in terms of visual object segmentation throughout grasp execution, we conduct grasping experiments with a set of five different objects during which we record the image data stream of all five in-finger cameras. The object set includes the four objects bowl, lemon, pitcher and strawberry from the  YCB Object Set \cite{doi:10.1177/0278364917700714} and additionally a green plastic cup. Each object is grasped in eleven trials where the grasp is controlled by an operator using an hand-attached shaft and the three button interface, shown in Fig. \ref{fig:graspingexperiment}.

\subsection{Perception Evaluation}
\begin{figure*}[h!]
	\centering

	\subfigure[]{
		\parbox{0.31\textwidth}{
			\label{fig:dg1}
			\includegraphics[trim={0cm 0 0cm 0cm},clip, width=0.31\textwidth]{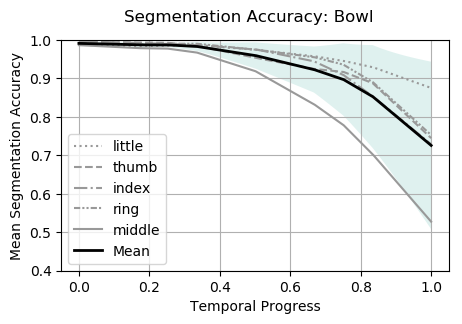}
	}}
	\subfigure[]{
		\parbox{0.31\textwidth}{
			\label{fig:dg2}
			\includegraphics[trim={0cm 0 0cm 0cm},clip, width=0.31\textwidth]{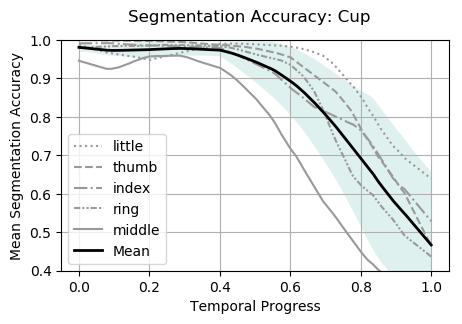}
	}}
	\subfigure[]{
		\parbox{0.31\textwidth}{
			\label{fig:dg3}
			\includegraphics[trim={0cm 0 0cm 0cm},clip, width=0.31\textwidth]{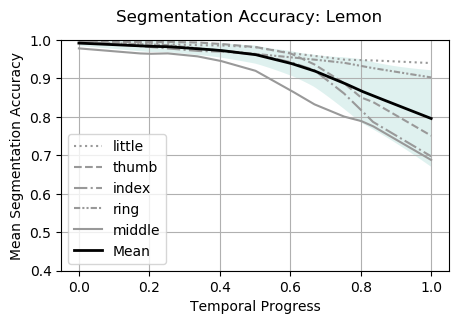}
	}}
	
	\subfigure[]{
		\parbox{0.31\textwidth}{
			\label{fig:dg4}
			\includegraphics[trim={0cm 0 0cm 0cm},clip, width=0.31\textwidth]{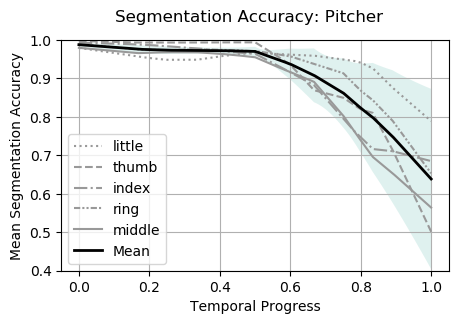}
	}}
	\subfigure[]{
		\parbox{0.3\textwidth}{
			\label{fig:dg5}
			\includegraphics[trim={0cm 0 0cm 0cm},clip, width=0.3\textwidth]{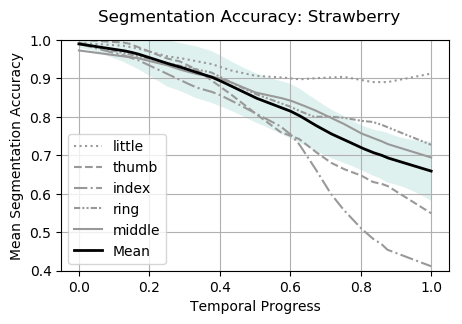}
	}}
	\subfigure[]{
		\parbox{0.3\textwidth}{
			\label{fig:quantSegBin}
			\includegraphics[trim={0cm 0 0cm 0cm},clip, width=0.3\textwidth]{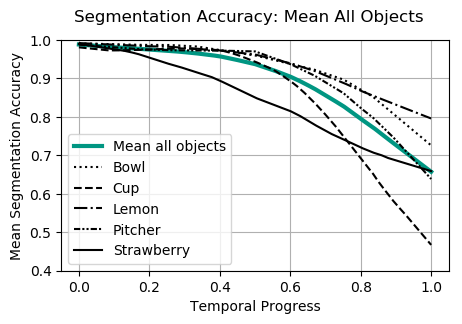}
	}}
	
	\caption{Evaluation of accuracy of the object segmentation during grasp experiments with five different test objects. The colored area indicates standard deviation.}
	
	\label{fig:AccEval}
	
\end{figure*}

We start data recording with the frame in which the image coverage of the object is minimal and stop at maximum image coverage. Each individual grasp execution (\textit{run}) is recorded with 6.47 frames in average that each includes five sub-images from the in-finger cameras. In total, a dataset including 1780 sub-images is recorded and annotated with binary ground-truth masks as shown in Fig. \ref{fig:segmenation}.
For faster data-transfer, the sub-images are down-sampled ($2\times 2$ filter) to a resolution of $88\times72\times3$ by the hand internal microprocessor before being transmitted to a PC.
We train and evaluate a class specific encoder-decoder CNN in a 11-fold cross validation where the data is divided into the 11 individual grasps trials. Captured and annotated data of one of the objects (bowl) was used as training and validation data set to optimize the hyperparameters. After hyperparameter optimization of the network architecture the remaining four objects are evaluated in an 11-fold cross validation.

We determine the accuracy of the segmentation which is expected to change over the course of the grasp progress. We evaluate the mean segmentation accuracy for all five tested objects per finger and calculate a mean accuracy value per object class depending on temporal progress of the grasp. Finally, we calculate the mean segmentation accuracy over all objects depending on temporal progress. Plots of these results can be found in Fig. \ref{fig:AccEval}. We obtain the mean accuracy of the classification over all grasps and object classes as $0.98 \percent$, $0.96\percent$, $0.89\percent$ and $0.74\percent$ in the 1\textsuperscript{st} to 4\textsuperscript{th} quartile of the temporal progress.

While in the first quartiles, good accuracies of more than 90~$\percent$ can be obtained, the 4\textsuperscript{th} quartile shows a significant drop in accuracy. This can be observed to varying degrees for all objects. As reasons for these inaccuracies, the decreasing camera quality and false image colors with smaller object distance can be named. An example of altered colors can be seen in Fig. \ref{fig:segmenation} in the bottom row image at 100~$\percent$.
To solve this problem, internal gain control or automatic color balance could be disabled. Also additional proximity information could be helpful.

\section{Conclusion}
In this work, we presented a soft humanoid hand that includes cameras for visual perception inside the fingertips. We propose a design of soft fingers that allows the mechatronic integration of cameras as the visual sensor system. The 3D-printed rigid palm includes three actuators in combination with an underactuated mechanism as well as a hybrid embedded system. The system allows processing of the multiple high data rate streams of visual information by using reconfigurable logic in combination with a microprocessor.
We evaluate the performance of the hand in terms of forces, grasp functionality as well as mechanical durability. The hand can exert grasp forces of up to \SI{11.6}{\newton} per finger and \SI{31.8}{\newton} in a cylindrical grasp. For individual finger actuation, we achieve nearly 5.000 closing cycles without any damage of the electrical connection and more than 15.000 actuation cycles before mechanical failure.
The presented hand can be used as a robotic hand but can also be used as a prosthetic hand prototype.

We designed an encoder-decoder network for object segmentation inside the camera images. The network provides pixel-wise semantic segmentation of objects during grasp process. At the beginning of the grasp process, mean accuracies of more than 90~$\percent$ can be achieved. Throughout the temporal progress of the grasp, the accuracy drops continuously. We see the reason in the camera internal image correction in challenging lighting conditions. This could be mitigated by a improved camera sensor or adding a sensor for obtaining additional depth information.

In future work we intend to realize hand internal image processing and scene interpretation using hardware acceleration. This will allow local extraction of relevant scene information and thereby significantly reduces the need for high bandwidth data connection to external processing units.
\addtolength{\textheight}{-6cm}   % This command serves to balance the column lengths
We see the new hand design with a set of multiple cameras located at potential points of contact as a basis for new kinematic control strategies of reactive grasping.
The evaluation of the new hardware design provides promising results and the possibility for obtaining visual feedback from cameras in the fingertip for more robust vision-based grasping.

\bibliographystyle{ieeetr}
\bibliography{citation}

\begin{thebibliography}{10}

\bibitem{hutchinson1996tutorial}
S.~Hutchinson, G.~D. Hager, and P.~I. Corke, ``A tutorial on visual servo
  control,'' {\em IEEE transactions on robotics and automation}, vol.~12,
  no.~5, pp.~651--670, 1996.

\bibitem{piazza2019century}
C.~Piazza, G.~Grioli, M.~Catalano, and A.~Bicchi, ``A century of robotic
  hands,'' {\em Annual Review of Control, Robotics, and Autonomous Systems},
  vol.~2, pp.~1--32, 2019.

\bibitem{Rus2015}
D.~Rus and M.~T. Tolley, ``{Design , fabrication and control of soft robots},''
  {\em Nature}, vol.~521, pp.~467--475, 2015.

\bibitem{Laschi2016}
C.~Laschi, B.~Mazzolai, and M.~Cianchetti, ``{Soft robotics: Technologies and
  systems pushing the boundaries of robot abilities},'' {\em Science Robotics},
  vol.~1, no.~1, 2016.

\bibitem{Gaiser2008}
I.~Gaiser, S.~Schulz, A.~Kargov, H.~Klosek, A.~Bierbaum, C.~Pylatiuk,
  R.~Oberle, T.~Werner, T.~Asfour, G.~Bretthauer, and R.~Dillmann, ``A new
  anthropomorphic robotic hand,'' in {\em IEEE/RAS International Conference on
  Humanoid Robots (Humanoids)}, pp.~418--422, 2008.

\bibitem{Deimel2016}
R.~Deimel and O.~Brock, ``{A novel type of compliant and underactuated robotic
  hand for dexterous grasping},'' {\em The International Journal of Robotics
  Research}, vol.~35, no.~1--3, pp.~161--185, 2016.

\bibitem{Tian2017}
M.~Tian, Y.~Xiao, X.~Wang, J.~Chen, and W.~Zhao, ``{Design and Experimental
  Research of Pneumatic Soft Humanoid Robot Hand},'' in {\em Robot Intelligence
  Technology and Applications 4}, pp.~469--478, 2017.

\bibitem{odhner2014compliant}
L.~U. Odhner, L.~P. Jentoft, M.~R. Claffee, N.~Corson, Y.~Tenzer, R.~R. Ma,
  M.~Buehler, R.~Kohout, R.~D. Howe, and A.~M. Dollar, ``A compliant,
  underactuated hand for robust manipulation,'' {\em The International Journal
  of Robotics Research}, vol.~33, no.~5, pp.~736--752, 2014.

\bibitem{Catalano2014}
M.~Catalano, G.~Grioli, E.~Farnioli, A.~Serio, C.~Piazza, and A.~Bicchi,
  ``{Adaptive synergies for the design and control of the Pisa/IIT SoftHand},''
  {\em The International Journal of Robotics Research}, vol.~33, no.~5,
  pp.~768--782, 2014.

\bibitem{Melchiorri2013}
C.~Melchiorri, G.~Palli, G.~Berselli, and G.~Vassura, ``{Development of the UB
  hand IV: Overview of design solutions and enabling technologies},'' {\em IEEE
  Robotics and Automation Magazine}, vol.~20, no.~3, pp.~72--81, 2013.

\bibitem{saudabayev2015sensors}
A.~Saudabayev and H.~A. Varol, ``Sensors for robotic hands: A survey of state
  of the art,'' {\em IEEE Access}, vol.~3, pp.~1765--1782, 2015.

\bibitem{hasegawa2010development}
H.~Hasegawa, Y.~Mizoguchi, K.~Tadakuma, A.~Ming, M.~Ishikawa, and M.~Shimojo,
  ``Development of intelligent robot hand using proximity, contact and slip
  sensing,'' in {\em 2010 IEEE International Conference on Robotics and
  Automation}, pp.~777--784, IEEE, 2010.

\bibitem{knoop2013dual}
E.~Knoop and J.~Rossiter, ``Dual-mode compliant optical tactile sensor,'' in
  {\em 2013 IEEE International Conference on Robotics and Automation},
  pp.~1006--1011, IEEE, 2013.

\bibitem{li2014GelSight}
R.~Li, R.~Platt, W.~Yuan, A.~ten Pas, N.~Roscup, M.~A. Srinivasan, and
  E.~Adelson, ``Localization and manipulation of small parts using gelsight
  tactile sensing,'' in {\em 2014 IEEE/RSJ International Conference on
  Intelligent Robots and Systems}, pp.~3988--3993, IEEE, 2014.

\bibitem{zhang2019vision}
S.~Zhang, J.~Shan, B.~Fang, F.~Sun, and H.~Liu, ``Vision-based tactile
  perception for soft robotic hand,'' in {\em 2019 IEEE International
  Conference on Robotics and Biomimetics (ROBIO)}, pp.~621--628, IEEE, 2019.

\bibitem{shimonomura2019tactile}
K.~Shimonomura, ``Tactile image sensors employing camera: A review,'' {\em
  Sensors}, vol.~19, no.~18, p.~3933, 2019.

\bibitem{balek1985using}
D.~Balek and R.~Kelley, ``Using gripper mounted infrared proximity sensors for
  robot feedback control,'' in {\em Proceedings. 1985 IEEE International
  Conference on Robotics and Automation}, vol.~2, pp.~282--287, IEEE, 1985.

\bibitem{yamaguchi2018gripper}
N.~Yamaguchi, S.~Hasegawa, K.~Okada, and M.~Inaba, ``A gripper for object
  search and grasp through proximity sensing,'' in {\em 2018 IEEE/RSJ
  International Conference on Intelligent Robots and Systems (IROS)}, pp.~1--9,
  IEEE, 2018.

\bibitem{hsiao2009reactive}
K.~Hsiao, P.~Nangeroni, M.~Huber, A.~Saxena, and A.~Y. Ng, ``Reactive grasping
  using optical proximity sensors,'' in {\em 2009 IEEE International Conference
  on Robotics and Automation}, pp.~2098--2105, IEEE, 2009.

\bibitem{yamaguchi2016}
A.~Yamaguchi and C.~G. Atkeson, ``Combining finger vision and optical tactile
  sensing: Reducing and handling errors while cutting vegetables,'' in {\em
  2016 IEEE-RAS 16th International Conference on Humanoid Robots (Humanoids)},
  pp.~1045--1051, IEEE, 2016.

\bibitem{shimonomura2016robotic}
K.~Shimonomura, H.~Nakashima, and K.~Nozu, ``Robotic grasp control with
  high-resolution combined tactile and proximity sensing,'' in {\em 2016 IEEE
  International Conference on Robotics and automation (ICRA)}, pp.~138--143,
  IEEE, 2016.

\bibitem{Weiner2018}
P.~Weiner, J.~Starke, F.~Hundhausen, J.~Beil, and T.~Asfour, ``{The KIT
  Prosthetic Hand: Design and Control},'' in {\em IEEE/RSJ Int. Conf. on
  Intelligent Robots and Systems}, 2018.

\bibitem{hundhausenresource}
F.~Hundhausen, D.~Megerle, and T.~Asfour, ``Resource-aware object
  classification and segmentation for semi-autonomous grasping with prosthetic
  hands,'' 2019.

\bibitem{lampe2013acquiring}
T.~Lampe and M.~Riedmiller, ``Acquiring visual servoing reaching and grasping
  skills using neural reinforcement learning,'' in {\em The 2013 international
  joint conference on neural networks (IJCNN)}, pp.~1--8, IEEE, 2013.

\bibitem{yan2017simtoreal}
M.~Yan, I.~Frosio, S.~Tyree, and J.~Kautz, ``Sim-to-real transfer of accurate
  grasping with eye-in-hand observations and continuous control,'' 2017.

\bibitem{viereck2017learning}
U.~Viereck, A.~t. Pas, K.~Saenko, and R.~Platt, ``Learning a visuomotor
  controller for real world robotic grasping using simulated depth images,''
  {\em arXiv preprint arXiv:1706.04652}, 2017.

\bibitem{morrison2018closing}
D.~Morrison, P.~Corke, and J.~Leitner, ``Closing the loop for robotic grasping:
  A real-time, generative grasp synthesis approach,'' {\em arXiv preprint
  arXiv:1804.05172}, 2018.

\bibitem{grabcad}
``{Eric Chen: R-Hand, Grabcad}.'' \url{https://grabcad.com/library/r-hand}.
\newblock Accessed: 2019-08-08.

\bibitem{fukaya2000design}
N.~Fukaya, S.~Toyama, T.~Asfour, and R.~Dillmann, ``Design of the
  tuat/karlsruhe humanoid hand,'' in {\em Proceedings. 2000 IEEE/RSJ
  International Conference on Intelligent Robots and Systems (IROS 2000)(Cat.
  No. 00CH37113)}, vol.~3, pp.~1754--1759, IEEE, 2000.

\bibitem{ronneberger2015unet}
O.~Ronneberger, P.~Fischer, and T.~Brox, ``U-net: Convolutional networks for
  biomedical image segmentation,'' 2015.

\bibitem{badrinarayanan2015segnet}
V.~Badrinarayanan, A.~Kendall, and R.~Cipolla, ``Segnet: A deep convolutional
  encoder-decoder architecture for image segmentation,'' 2015.

\bibitem{doi:10.1177/0278364917700714}
B.~Calli, A.~Singh, J.~Bruce, A.~Walsman, K.~Konolige, S.~Srinivasa, P.~Abbeel,
  and A.~M. Dollar, ``Yale-cmu-berkeley dataset for robotic manipulation
  research,'' {\em The International Journal of Robotics Research}, vol.~36,
  no.~3, pp.~261--268, 2017.

\end{thebibliography}


%\IEEEtriggeratref{10}

\end{document}